\documentclass{article}
\usepackage{ijcai23}

\usepackage{times}
\usepackage{soul}
\usepackage{url}
\usepackage[hidelinks]{hyperref}
\usepackage[utf8]{inputenc}
\usepackage[small]{caption}
\usepackage{graphicx}
\usepackage{amsmath}
\usepackage{amsthm}
\usepackage{booktabs}
\usepackage{algorithm}
\usepackage{algorithmic}
\usepackage[switch]{lineno}
\usepackage{adjustbox}

\usepackage{amsmath,amsfonts,bm}

\def\eqref#1{equation~\ref{#1}}

\def\1{\bm{1}}

\def\vh{{\bm{h}}}

\def\vx{{\bm{x}}}

\def\vz{{\bm{z}}}

\def\mA{{\bm{A}}}
\def\mB{{\bm{B}}}

\def\mH{{\bm{H}}}

\def\mK{{\bm{K}}}

\def\mN{{\bm{N}}}

\def\mQ{{\bm{Q}}}

\def\mV{{\bm{V}}}

\def\mX{{\bm{X}}}

\def\mZ{{\bm{Z}}}

\DeclareMathAlphabet{\mathsfit}{\encodingdefault}{\sfdefault}{m}{sl}
\SetMathAlphabet{\mathsfit}{bold}{\encodingdefault}{\sfdefault}{bx}{n}

\def\gN{{\mathcal{N}}}

\def\gR{{\mathcal{R}}}
\def\gS{{\mathcal{S}}}

\newcommand{\R}{\mathbb{R}}

\usepackage{multirow}
\usepackage{multicol}

\title{Hierarchical Transformer for Scalable Graph Learning}

\author{
Wenhao Zhu$^1$
\and
Tianyu Wen$^2$\and
Guojie Song$^1$\and
Xiaojun Ma$^3$\And
Liang Wang$^4$
\affiliations
$^1$National Key Laboratory of General Artificial Intelligence, School of Intelligence Science and Technology, Peking University\\
$^2$Yuanpei College, Peking University\\
$^3$Microsoft  $^4$Alibaba Group
\emails
\{wenhaozhu, tianyuwen, gjsong\}@pku.edu.cn,
xiaojunma@microsoft.com,
liangbo.wl@alibaba-inc.com
}

\makeatletter
\def\blfootnote{\xdef\@thefnmark{}\@footnotetext}
\makeatother

\begin{document}

\maketitle

\begin{abstract}
    Graph Transformer is gaining increasing attention in the field of machine learning and has demonstrated state-of-the-art performance on benchmarks for graph representation learning. However, as current implementations of Graph Transformer primarily focus on learning representations of small-scale graphs, the quadratic complexity of the global self-attention mechanism presents a challenge for full-batch training when applied to larger graphs. Additionally, conventional sampling-based methods fail to capture necessary high-level contextual information, resulting in a significant loss of performance. In this paper, we introduce the Hierarchical Scalable Graph Transformer (HSGT) as a solution to these challenges. HSGT successfully scales the Transformer architecture to node representation learning tasks on large-scale graphs, while maintaining high performance. By utilizing graph hierarchies constructed through coarsening techniques, HSGT efficiently updates and stores multi-scale information in node embeddings at different levels. Together with sampling-based training methods, HSGT effectively captures and aggregates multi-level information on the hierarchical graph using only Transformer blocks. Empirical evaluations demonstrate that HSGT achieves state-of-the-art performance on large-scale benchmarks with graphs containing millions of nodes with high efficiency.
\end{abstract}

\vspace{-3mm}

\section{Introduction}

Transformer \cite{vaswani2017attention} is now the prevalent universal neural architecture in natural language processing and computer vision with its powerful, lowly inductive-biased self-attention mechanism. The great success of Transformer has encouraged researchers to explore its adaptation to graph machine learning on node-level and graph-level tasks \cite{ying2021transformers,kreuzer2021rethinking,chen2022structure}. While GNNs are known to suffer from inherent limitations in the message-passing paradigm like \textit{over-smoothing} and \textit{neighbor explosion}, the promising performance of these graph Transformer methods has encouraged researchers to expand the Transformer architecture to more scenarios in graph machine learning.

Still, challenges arise when scaling Transformer to large graphs. In previous graph Transformer methods, self-attention calculates all pairwise interactions in a graph, indicating that it has quadratic complexity to the total number of nodes. Thus, to perform training on graphs of millions of nodes without substantial modification to the Transformer architecture, one must sample a properly sized subgraph at every batch so that the computational graph can be fit into GPU memory. Using sampling strategies like neighbor sampling in GraphSAGE \cite{hamilton2018inductive}, we can build a simple scalable graph Transformer model by directly applying existing models like Graphormer on the sampled subgraph. However, there is an intrinsic weakness in this straightforward combination of Transformer architecture and sampling-based training methods. It has been widely observed that high-level context information characterized by global receptive field of self-attention module greatly contributes to Transformer's outstanding performance \cite{vaswani2017attention,dai2019transformer,liu2021swin,ying2021transformers}. Considering that the entire input graph is usually far larger than every sampled subgraph, when the receptive field of each node is restricted to the sampled local context, the model may ignore high-level context information, leading to possible performance loss. Meanwhile, if we add globally sampled nodes to the sampled set to reduce context locality, it is likely to introduce much redundant noise because most long-distance neighbors are irrelevant in large graphs. This hypothesis is further confirmed by our experiments.

\begin{figure}
\begin{center}
\includegraphics[width=0.7\linewidth]{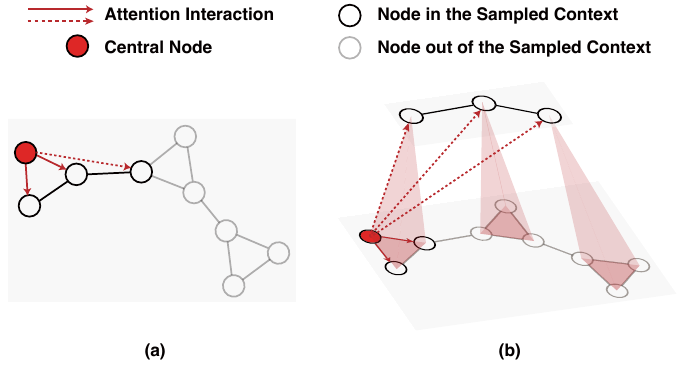}
\end{center}
\caption{(a) During straightforward sampling-based training, receptive field of each node is restricted to the sampled context, leading to the loss of global context information. (b) In our method, through attention interaction with nodes at higher hierarchies, the receptive field of each node is expanded to a higher scale (red shadows), making it possible for the model to capture high-level knowledge.}
\label{fig_intro}
\vspace{-5mm}
\end{figure}

In this paper, to alleviate the problem above and find the real potentials of Transformer architecture on large-scale graph learning tasks, we propose HSGT, a hierarchical scalable graph Transformer framework. Our key insight is that, by building graph hierarchies with topological coarsening methods, high-level context information can be efficiently stored and updated with a fused representation of high-level node embeddings. As illustrated in Figure \ref{fig_intro}, through attention interaction with nodes at higher hierarchical layers, the receptive field of each node is expanded to a much higher scale, making it possible for HSGT to effectively capture high-level structural knowledge in the graph during sampling-based training. Since the number of high-level nodes is marginal compared to the size of original graph, our approach is efficient and brings low extra computational cost. Besides, using adaptively aggregated representation to characterize high-level context information, our method is robust to random noise from long-range irrelevant neighbors.

More concretely, our proposed HSGT architecture utilizes three types of Transformer blocks to support context transformations at different scales. At every hierarchical layer, the horizontal blocks are first performed to exchange and transform information in the local context, then we use vertical blocks to aggregate representation in every substructure and create embeddings for nodes at the higher level. Eventually, readout blocks obtain final node representation by fusing multi-level node embeddings. To achieve scalable training, we have developed a hierarchical sampling method to sample multi-level batches for training and inference, and utilized the historical embedding technique \cite{fey2021gnnautoscale} to remove inter-batch dependencies and prune the computational graph. Being completely Transformer-based, the resulting HSGT architecture is highly scalable and generalizable, achieving state-of-the-art results on a wide range of datasets from the standard Cora \cite{sen2008collective} to ogbn-products \cite{chiang2019cluster} with millions of nodes, outperforming the standard scalable GNN and Transformer baselines. We summarize our main contributions as follows:
\begin{itemize}
    \item We propose HSGT, a new graph Transformer architecture that efficiently generates high-quality representations for graphs of varying sizes via effective usage of graph hierarchical structure and multi-level network design.
    \item We develop the novel hierarchical sampling strategies and apply the historical embedding method, which allow HSGT to be trained efficiently large-scale graphs and gain high performance.
    \item Extensive experiments show that HSGT achieves state-of-the-art performance against baseline methods on large-scale graph benchmarks with computational costs similar to the standard GraphSAGE method. 
\end{itemize}

\section{Related Work}
\subsection{Graph Transformers}

Along with the recent surge of Transformer, many prior works have attempted to bring Transformer architecture to the graph domain, including GT \cite{dwivedi2020generalization}, GROVER \cite{rong2020self}, Graphormer \cite{ying2021transformers}, SAN \cite{kreuzer2021rethinking}, SAT \cite{chen2022structure}, ANS-GT \cite{zhang2022hierarchical}, GraphGPS \cite{rampavsek2022recipe} and NodeFormer \cite{wunodeformer}. Graphormer \cite{ying2021transformers} proposes an enhanced Transformer with centrality, spatial and edge encodings, and achieves state-of-the-art performance on many molecular graph representation learning benchmarks. SAN \cite{kreuzer2021rethinking} presents a learned positional encoding that cooperates with full Laplacian spectrum to learn the position of each node in the graph. Gophormer \cite{zhao2021gophormer} applies structural-enhanced Transformer to sampled ego-graphs to improve node classification performance and scalability. SAT \cite{chen2022structure} studies the question of how to encode structural information to Transformers and proposes the Structure-Aware-Transformer to generate position-aware information for graph data. ANS-GT \cite{zhang2022hierarchical} proposes an adaptive sampling strategy to effectively scale up graph Transformer to large graphs. GraphGPS \cite{rampavsek2022recipe} proposes a general, powerful and scalable architecture for building graph Transformers and use the fast attention method to improve scalability. NodeFormer \cite{wunodeformer} introduces a novel all-pair message passing
scheme for efficiently propagating node signals between arbitrary nodes based on the Gumbel-Softmax operator.

\subsection{Scalable Graph Learning}
Modeling complex, real-world graphs with large scale require scalable graph neural models. On large graphs, message-passing GNNs mainly suffer from the \textit{neighbor explosion} phenomenon, since the neighborhood dependency of nodes grows exponentially as the model depth increases, which results in the excessive expansion of computational graphs. Sampling-based methods \cite{hamilton2018inductive,chen2018fastgcn,chen2017stochastic,chiang2019cluster,zeng2019graphsaint,huang2018adaptive} generally solve this issue by running model on the sampled subgraph batches, and offline propagation methods \cite{wu2019simplifying,klicpera2018predict,frasca2020sign,zhang2021graph} achieve fast training and inference by decoupling feature propagation from prediction as a pre-processing step. Notably, historical embedding methods \cite{chen2017stochastic,fey2021gnnautoscale} store intermediate node embeddings from previous training iterations and use them as approximations for accurate embeddings.

\section{Preliminaries}
In this section we present some model backgrounds and basic notations. Let $G=(\mathcal V,\mathcal E)$ denote a graph, where $\mathcal V=\{v_1,v_2,\ldots,v_n\}$ is the node set that consists of $n$ vertices and $\mathcal E\subset \mathcal V\times \mathcal V$ is the edge set. For node $v\in\mathcal V$, let $\mathcal N(v)=\{v':v'\in\mathcal V, (v,v')\in\mathcal E\}$ denote the set of its neighbors.  Let each node $v_i$ be associated with a feature vector $\vx_i\in \mathbb{R}^F$ where $F$ is the hidden dimension, and let $\mathbf{X}=[\vx_1,\vx_2,\ldots,\vx_n]^\top\in\mathbb{R}^{n\times F}$ denote the feature matrix.
\subsection{Transformer}

\paragraph{Standard Transformer Layers.} Transformer \cite{vaswani2017attention} is first proposed to model sequential text data with consecutive Transformer layers, each of which mainly consists of a multi-head self-attention (MHA) module and a position-wise feed-forward network (FFN) with residual connections. For queries $\mQ\in\R^{n_q\times d}$, keys $\mK\in\R^{n_k\times d}$ and values $\mV\in\R^{n_k\times d}$, the scaled dot-product attention module can be defined as
\begin{align}
    \text{Attention}(\mQ,\mK,\mV)=\text{softmax}(\mA)\mV,\mA=\frac{\mQ\mK^{\top}}{\sqrt{d}},
\end{align}
where $n_q, n_k$ are number of elements in queries and keys, and $d$ is the hidden dimension. 
After multi-head attention, the position-wise feed-forward network and layer normalization are performed on the output.

\paragraph{Biased Transformer Layers.} A bias term can be added to the attention weights $\mA$ to represent pair-wise knowledge like relative positional encodings in \cite{shaw2018self}. Suppose we have a bias matrix $\mB\in\R^{n_q\times n_k}$, the biased-MHA can be formulated by replacing the standard attention module with the attention weight matrix computed by $\mA=\frac{\mQ\mK^{\top}}{\sqrt{d}}+\mB.$

\begin{figure*}
\begin{center}
\includegraphics[width=0.9\linewidth]{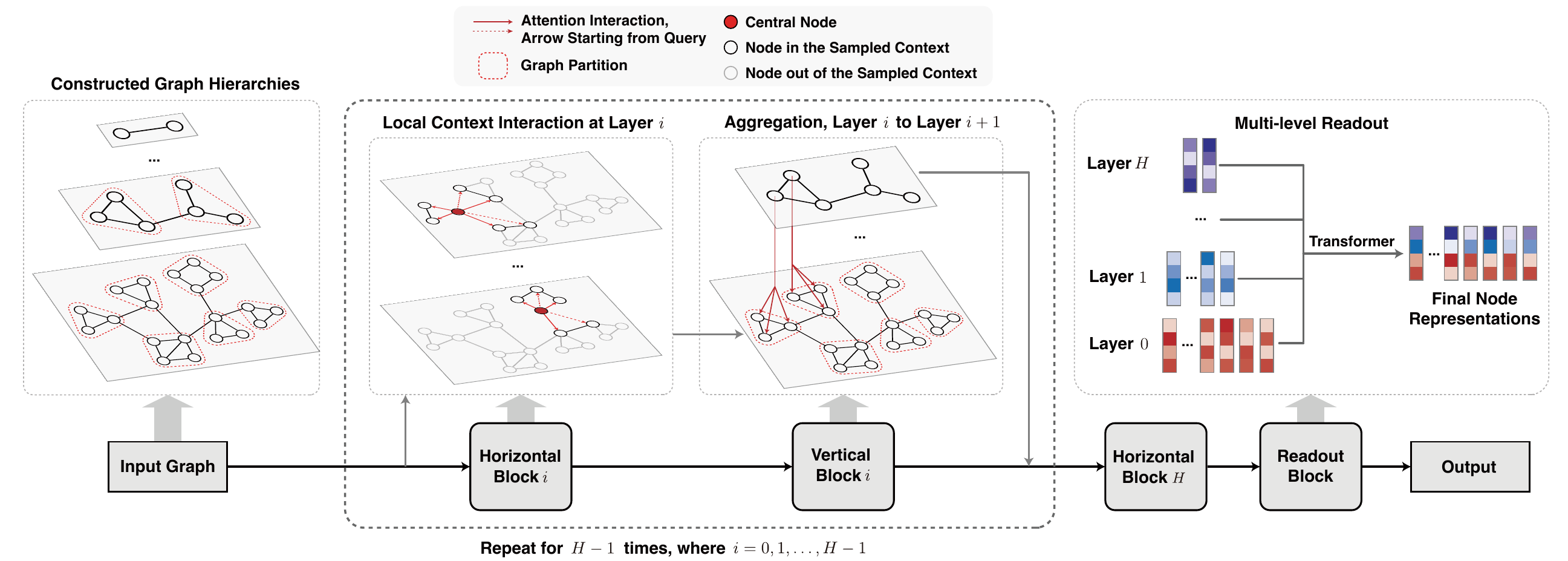}
\end{center}
\caption{A high-level illustration of the proposed HSGT model architecture. At every hierarchical layer, a horizontal block first exchanges and transforms information in each node's local context, then a vertical block is performed to adaptively coalesce every substructure if a higher layer exists. At last, a readout block aggregates multi-level representations to calculate the final output. Details are omitted in this illustration for clarity.}
\vspace{-5mm}
\label{fig_overview}
\end{figure*}

\subsection{Graph Hierarchies}
For graph $G^0=(V^0,E^0)$, graph coarsening aims to find $G^1=(V^1,E^1)$ that captures the essential substructures of $G_0$ and is significantly smaller ($|V^1|\ll |V^0|$,$|E^1|\ll |E^0|$). We assume graph coarsening is performed by coalescing nodes with a surjective mapping function $\phi:V^0\to V^1$. Every node $v_i^1\in V^1$ corresponds to a node cluster $\phi^{-1}(v_i^1)=\{v_j^0\in V^0:\phi(v_j^0)=v_i^1\}$ in $G^0$, and the edge set of $G^1$ is defined as $E^1=\{(v_i^1,v_j^1):\exists v_r^0\in\phi^{-1}(v_i^1),v_s^0\in\phi^{-1}(v_j^1),\text{such that }(v_r^0,v_s^0)\in E^0\}$. We also initialize node embeddings of $G^1$ by $\vx_i^1=\text{Mean}(\{\vx_j^0:v_j^0\in\phi^{-1}(v_i^1)\})$ for every node $v_i^1$ in $V^1$. The coarsening ratio $\alpha$ at this step is defined as $\alpha=\frac{|V^1|}{|V^0|}$. By running the coarsening algorithm recursively, a graph hierarchy $\{G^0,G^1,$ $\ldots,G^H\}$ can be constructed to summarize multi-level structures.

\section{Proposed Approach}
In the following section, we will describe the motivation and approach behind the creation of our HSGT model, and subsequently provide a detailed description of the architecture of the entire model. Figure \ref{fig_overview} gives a high-level illustration of model architecture, and Algorithm \ref{alg1} describes the computation steps of HSGT.

\subsection{Motivation for Utilizing Graph Hierarchies}

The key difference between HSGT and previous graph Transformers is the utilization of graph hierarchical structure. Previous methods for leveraging high-level context information can only expand the receptive field and increase sample size, which leads to significant computational overhead and performance loss on node-level tasks. In contrast, HSGT utilizes graph hierarchies to communicate high-level information via the embedding of virtual nodes, resulting in several key benefits: (1) low computational cost: the number of high-level virtual nodes is minimal and the updating process is completed via efficient sampling strategies and historical embedding; (2) broad receptive field: the receptive field of a single node includes any node related to its corresponding high-level nodes; (3) high flexibility: one can build the graph hierarchy using any graph coarsening algorithm, and arbitrarily choose the sampling strategy to control the structural information involved in the learning process. In the following paragraphs, we will elaborate on the design of the model and demonstrate its effectiveness through experiments.

\subsection{Model Architecture}

\begin{algorithm}[h]
    \caption{Overview of HSGT}
    \label{alg1}
    \textbf{Input}: Input graph $G=(V,E)$, with corresponding hierarchy $\{G^0,G^1,$ $\ldots,G^H\}$, initial feature matrices $\mX_0,\mX_1,\ldots,\mX_H$ and hierarchical mappings $\phi_1,\ldots,\phi_H$, batch size $B$. \\
    \textbf{Output}: embedding $\vh_v$ for every node $v$. \\
    \vspace{-4mm}
    \begin{algorithmic}[1] 
        \STATE $V^H_{\text{left}}\leftarrow V^H$, where $G^H=(V^H,E^H)$.
        \WHILE{$V^H_{\text{left}}$ is not empty}
        \STATE Sample $V^H_B$ from $V^H_{\text{left}}$ with $|V^H_B|=B$.
        \STATE $V^H_{\text{left}}\leftarrow V^H\setminus V^H_B$.
        \FOR{$j$ in $1,2,\ldots,H$}
        \STATE $\tilde{V^{j-1}_B}\leftarrow\phi^{-1}_j(V^j_B)$.
        \STATE $V^{j-1}_B\leftarrow \text{NeighborSample}(\tilde{V^{j-1}_B})$. (Section \ref{sec33})
        \STATE $\mH_{j-1}\leftarrow \mX_{j-1}[V^{j-1}_B]$.
        \ENDFOR
        \FOR{$j$ in $0,1,2,\ldots,H$}
        \STATE $\mH_j\leftarrow \text{HorizontalBlock}(\mH_j)$.
        \IF{$j<H$}
        \STATE $\mH_{j+1}\leftarrow \text{VerticalBlock}(\mH_j,\mH_{j+1})$.
        \ENDIF
        \ENDFOR
        \STATE $\mH_0\leftarrow \text{ReadoutBlock}(\mH_0,\mH_1,\ldots,\mH_H)$.
        \STATE $\vh_v\leftarrow \mH_0[v], \forall v\in V_B^H$.
        \ENDWHILE
        \STATE \textbf{return} $\vh_v, \forall v\in V$.
    \end{algorithmic}
\end{algorithm}

\subsubsection{Graph Hierarchy Construction and Input Transformation}

For input graph $G^0=(V^0,E^0)$ with feature matrix $\mX$, we first use the chosen coarsening algorithm to produce the graph hierarchy $\{G^0,G^1,$ $\ldots,G^H\}$ with initial feature matrices $\mX_0,\mX_1,\ldots,\mX_H$ and corresponding hierarchical mappings $\phi_1,\ldots,\phi_H$. The number of hierarchical layers $H$ and the coarsening ratios for each step $\alpha_1,\ldots,\alpha_H$ are pre-defined as hyperparameters. The model imposes no limitations on the specific graph coarsening algorithm. In our implementation, we choose \textsc{METIS} \cite{karypis1997metis} which is designed to partition a graph into mutually exclusive groups and minimize the frequency of inter-links between different groups.  The chosen modern METIS algorithm is fast and highly scalable with time complexity approximately bounded by $O(|E|)$, and only needs to compute once at the pre-processing stage. In experiments, even partitioning of the largest \textit{ogbn-products} graph is finished within 5 minutes, bringing almost no overhead to the entire training process.

We also apply linear transformations and degree encodings for initial features before they are fed into Transformer blocks. All feature vectors in $\mX_0,\mX_1,\ldots,\mX_H$ are projected into $\R^d$ using a linear layer, where $d$ is the hidden size. Additionally, inspired by centrality encoding in \cite{ying2021transformers}, we add learnable embedding vectors indexed by node degree to the transformed node feature at layer 0 to represent structural information.

\subsubsection{Horizontal Block}

In every horizontal block, we aim to horizontally aggregate and transform node representations in every node's local context using structural-enhanced Transformer layers. Suppose the input graph is $G=(V,E)$ with $n$ nodes, the feature matrix is $\mH$, and every node $v$ has an individual local receptive field $\gR(v)\subset V$ which we will discuss later. Following \cite{ying2021transformers}, to leverage graph structure into self-attention, we choose to quantify the connectivity between nodes with the distance of the shortest path (SPD), and use an attention bias term to represent the structural information. To reduce the computational cost, we set a maximum SPD length $D$ such that SPDs longer than $D$ will not be computed. We also mask nodes not in $\gR(v)$ out of the receptive field of $v$ by setting the corresponding bias term to $-\infty$. Formally, the bias matrix $\mB\in\R^{n\times n}$ is defined as
\begin{align*}
    \mB_{i,j}=
    \begin{cases}
        b_{\text{SPD}(v_i,v_j)},&\text{if } v_j\in\gR(v_i) \text{, SPD}(v_i,v_j)\leq D,\\
        -\infty,&\text{if }v_j\notin\gR(v_i),\\
        0,&\text{else.}
    \end{cases}
\end{align*}
where $\mB_{i,j}$ is the $(i,j)$-element of $\mB$, $b_0,b_1,\ldots,b_D\in\R$ are a series of learnable scalars. Then the horizontal block is built by stacking the following Transfomer layers:
\begin{align}
    \mH=\text{Biased-Transformer}(\mH,\mH,\mH,\mB).
\end{align}

Compared with GNNs, our method promotes a high-order knowledge aggregation in a broader context while leveraging the complete structural information. In the case of full-batch training on small graphs, every node can have a global receptive field, i.e. $\gR(v)=V,\forall v \in V$. But during sampling-based training on huge graphs, the receptive field of each node is restricted to the current batch $V_B\subset V$, and we empirically discover that due to the unbalance and irregularity of sampling methods, computing every pair-wise attention in $V_B$ will lead to significant performance drop on node-level tasks. In the mean time, intra-batch communication will be limited if we set the receptive field of each node to its local neighbors. To balance the two aspects, we choose to form the receptive field $\gR(v)$ of node $v$ with its $D$-hop neighbors $\gN_D(v)$ and nodes individually randomly sampled from $V_B$ by probability $p$. Besides, during experiments we discover that by sharing parameters among horizontal blocks at different hierarchical layers, we can greatly reduce the number of model parameters while the model performance is not affected. This is probably due to the structural similarity between graphs at different hierarchical levels and the strong expressive capacity of Transformer layers. We will test the effectiveness of the two strategies in ablation studies.

\subsubsection{Vertical Block}

The vertical block focuses on aggregating node representations produced by the previous horizontal block and generating embeddings for nodes at next hierarchical level. In contrast to simple pooling functions like mean and sum, to overcome their incapability of capturing important nodes and substructures, we reuse the attention mechanism to adaptively merge vector embeddings. Suppose the vertical block aims to calculate embeddings for nodes in $G^{i+1}$. For node $v\in V^{i+1}$ with transformed initial feature $\vx_v$, its representation $\vh_v$ after the vertical aggregation is computed as
\begin{align}
    &\mN_v=\text{Stack}(\{\vh_s:s\in\phi_{i+1}^{-1}(v)\}),\\
    &\vh_v=\text{Transformer}(\vx_v,\mN_v,\mN_v),
\end{align}
where $\vx_v,\vh_v$ are viewed as matrices in $\R^{1\times d}$, and the vertical block computes representation for every node in the input. This aggregation scheme allows every fused representation $\vh_v$ to contain meaningful information on its corresponding low-level substructure, helping the next horizontal block to achieve better high-level knowledge exchange.

\subsubsection{Readout Block}

After the horizontal blocks are performed on every hierarchical level, we use the readout block to fuse representations and produce final node embeddings. This operation brings well aggregated multi-level information to nodes in the original graph, expanding their receptive field to a much higher scale. Formally, let $\vh_r$ denote the embedding of node $r$ at level $l$ generated by the $l$-th horizontal block, then for a node $v\in V^0$, its final output embedding $\overline{\vh}_v$ is calculated through attention with its corresponding high-level nodes:
\begin{align}
    &\mZ_v=[\vh_{t_0},\vh_{t_1},\ldots,\vh_{t_H}]^{\top},\\
    &\overline{\vh}_v=\text{Transformer}(\vh_v,\mZ_v,\mZ_v),
\end{align}
where $t_0=v,t_j=\phi_j(t_{j-1}),\text{for }j=1,\ldots,H$.

\subsection{Training Strategy}
\label{sec33}

\subsubsection{Hierarchical Sampling}
\begin{figure}
\begin{center}
\includegraphics[width=0.8\linewidth]{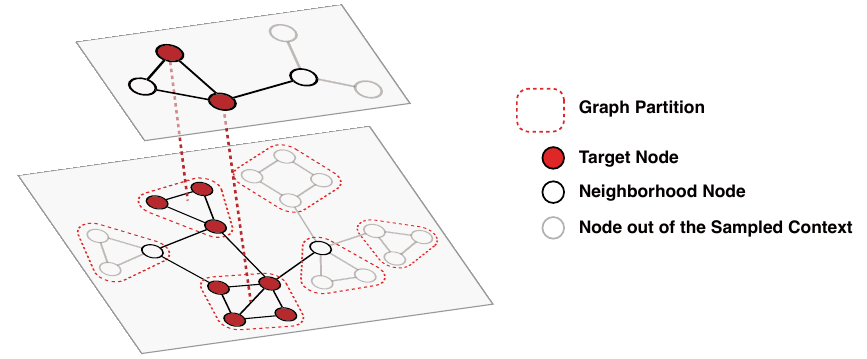}
\end{center}
\caption{An illustration of the proposed sampling method.}
\label{fig_sampling}
\vspace{-5mm}
\end{figure}

When training on large-scale graphs, at every step, our model can only be operated on a sampled batch $\{G^0_B,G^1_B,\ldots,G^H_B\}$ of the entire graph hierarchy $\{G^0,G^1,\ldots,G^H\}$, that $G^i_B$ is a subgraph of $G^i$ for every $i=0,1,\ldots,H$. To keep every substructure in low-level hierarchical layers complete, we follow a top-to-bottom sampling approach. For batch size $b$, we first randomly sample $b$ nodes at $V^H$ as $\tilde V_B^H$, then recursively sample nodes from layer $H-1$ to $0$ in $\tilde V^{i-1}_B=\bigcup_{v\in\tilde V_B^i}\phi_i^{-1}(v),i=H,H-1,\ldots,1. $
Here we call nodes in $\{\tilde V_B^0,\tilde V_B^1,\ldots,\tilde V_B^H\}$ \textit{target nodes}, because only representations of nodes in $\tilde V_B^0$ will be used for supervised learning and $\{\tilde V_B^1,\ldots,\tilde V_B^H\}$ contains all high-level nodes they relate to. Meanwhile, to promote local context interaction in horizontal blocks, we additionally sample a neighborhood set for every target node at all levels using neighbor sampling. An illustration of this process is presented in Figure \ref{fig_sampling}. We construct the final sampled nodes set $\{V_B^0,V_B^1,\ldots,V_B^H\}$ by adding the neighborhood sets to the target nodes set, and $\{G^0_B,G^1_B,\ldots,G^H_B\}$ are the corresponding induced subgraphs. The resulting sampling strategy allows the model to operate on complete hierarchical structures with local context preserved, which is critical for the horizontal and vertical modules to work well. 

\subsubsection{Historical Embeddings for High-level Nodes}

The sampling scheme above could cause issues. For example, for neighborhood node $v\in V_B^1\setminus\tilde V_B^1$, it is very likely that most of its corresponding nodes at layer 0 will not appear in the sampled nodes set $V_B^0$, since we do not deliberately add $\phi^{-1}_1(v)$ to the sampled set as the target nodes do. Thus, it is not possible to directly get the representation of $v$ by aggregating embeddings of nodes in $\phi^{-1}_1(v)$ via vertical blocks when some of $\phi^{-1}_1(v)$ do not exist in the sampled context. And manually adding those nodes to the data batch will lead to a massive expansion of the computational graph (almost $10\times$). Nevertheless, if we skip the neighborhood sampling step for high-level nodes, the inter-batch communication of structural knowledge could be baffled, which contradicts with our initial goal.

To disentangle such multi-level dependencies, we utilize the historical embedding method proposed in \cite{chen2017stochastic,fey2021gnnautoscale} to alleviate the \textit{neighbor explosion} in GNNs. In our model, the historical embeddings act as an offline storage $\gS$ of high-level nodes (above level $0$), which is accessed and updated at every batch using \textit{push} and \textit{pull} operations. At every batch, the vertical blocks are only performed on the target nodes, and we push the newly aggregated embeddings for high-level target nodes to $\gS$. For horizontal blocks on high-level nodes, we approximate the embeddings of neighborhood nodes via pulling historical embeddings in $\gS$ acquired in previous batches. Specifically, for horizontal block $i$, its input $\mH$ will be computed as
\begin{align}
    &\mH=\text{Stack}(\{\vh_s:s\in V^i_B\}),\\
    &\approx\text{Stack}(\{\vh_s:s\in\tilde V^i_B\}\cup\{\tilde \vh_s:s\in V^i_B\setminus\tilde V^i_B\}),
\end{align}
where $\vh_s$ denotes accurate embedding of $s$ calculated by the previous horizontal block, and $\tilde\vh_s$ denotes the historical embedding of $s$ from previous batches. With historical embeddings, we enable the inter-batch communication of high-level contexts with low extra computational cost and high accuracy bounded by theoretical results in \cite{chen2017stochastic,fey2021gnnautoscale}. Our approach is the first implementation of the historical embedding method on graph Transformer models, and its effectiveness is further demonstrated by the ablation studies.

\begin{table*}[htbp]
    \begin{adjustbox}{width=2.1\columnwidth,center}
    \begin{tabular}{l|cccc|cccccc}
        \toprule
        \#nodes & 2.7K & 3.3K & 19.7K & 7.6K & 169K & 233K & 89.2K & 716K & 133K & 2.4M \\
        \#edges & 5.2K & 4.5K & 44.3K & 119K & 1.1M & 11.6M & 450K & 7.0M & 40M & 61.8M \\
        Dataset & \textsc{Cora} & \textsc{CiteSeer} & \textsc{PubMed} & \textsc{Amazon-Photo} & \texttt{ogbn-arxiv} & \textsc{Reddit} & \textsc{Flickr} & \textsc{Yelp} & \texttt{ogbn-proteins} & \texttt{ogbn-products}\\
        \midrule
        GCN & 77.20\small$\pm 1.51$ & 69.49\small$\pm 0.58$ & 77.60\small$\pm 0.96$ & 92.44\small$\pm 0.22$ & 71.74\small$\pm 0.29$* & 91.01\small$\pm 0.29$ & 51.86\small$\pm 0.10$ & 32.14\small$\pm 0.66$ & 72.51\small$\pm 0.35$* & 75.64\small$\pm 0.21$* \\
        GAT & 82.80\small$\pm 0.47$ & 69.20\small$\pm 0.45$ & 76.90\small$\pm 0.85$ & 92.88\small$\pm 0.37$ & 57.88\small$\pm 0.18$ & 96.50\small$\pm 0.14$ & 52.39\small$\pm 0.05$ & 61.58\small$\pm 1.37$ & 72.02\small$\pm 0.44$* & 79.45\small$\pm 0.59$ \\
        GIN & 75.93\small$\pm 0.99$ & 63.83\small$\pm 0.49$ & 77.03\small$\pm 0.42$ & 80.14\small$\pm 1.46$ & 52.23\small$\pm 0.34$ & 86.37\small$\pm 0.62$ & 48.28\small$\pm 0.85$ & 29.75\small$\pm 0.86$ & 70.76\small$\pm 0.08$& 74.79\small$\pm 0.81$ \\
        GraphSAGE & - & - & - & - & 71.49\small$\pm 0.27$* & 96.53\small$\pm 0.11$ & 51.86\small$\pm 0.35$ & 53.89\small$\pm 0.85$ & 77.68\small$\pm 0.20$* & 78.50\small$\pm 0.14$*\\
        Cluster-GCN & - & - & - & - & 69.76\small$\pm 0.49$* & 95.12\small$\pm 0.08$ & 50.25\small$\pm 0.83$ & 52.50\small$\pm 0.19$ & 74.89\small$\pm 0.12$ & 78.97\small$\pm 0.33$* \\
        GraphSAINT & - & - & - & - & 58.63\small$\pm 0.33$ & 90.92\small$\pm 0.61$ & 51.91\small$\pm 0.06$ & 56.22\small$\pm 1.14$ & 70.22\small$\pm 0.84$ & 79.08\small$\pm 0.24$*\\
        GAS-GCN & 82.29\small$\pm 0.76$* & \textbf{71.18}\small$\pm 0.97$* & 79.23\small$\pm 0.62$* & 90.53\small$\pm 1.40$* & 71.68* & 95.45* & 54.00* & 62.94* & - & 76.66* \\
        SIGN & - & - & - & - & - & 96.8\small$\pm 0.0$* & 51.4\small$\pm 0.1$* & 63.1\small$\pm 0.3$* & - & 77.60\small$\pm 0.13$*\\
        GraphZoom & - & - & - & - & 71.18\small$\pm 0.18$* & 92.5* & - & - & - & 74.06\small$\pm0.26$*\\
        \midrule
        Graphormer & 66.35\small$\pm 2.44$ & 56.22\small$\pm 3.27$ & OOM & OOM & OOM & OOM & OOM & OOM & OOM & OOM\\
        Graphormer-\textsc{Sample} & 75.14\small$\pm 1.31$ & 61.46\small$\pm 1.90$ & 75.45\small$\pm 0.98$ & 92.76\small$\pm 0.59$ & 70.43\small$\pm 0.20$ & 93.05\small$\pm 0.22$ & 51.93\small$\pm 0.21$ & 60.01\small$\pm 0.45$ & 72.34\small$\pm 0.51$ & 79.10\small$\pm 0.12$ \\
        SAN & 36.61\small$\pm 3.49$ & 44.35\small$\pm1.08$ & OOM & OOM & OOM & OOM & OOM & OOM & OOM & OOM\\
        SAT & 72.40\small$\pm 0.31$ & 60.93\small$\pm 1.25$ & OOM & OOM & OOM & OOM & OOM & OOM & OOM & OOM\\
        SAT-\textsc{Sample} & 74.55\small$\pm 1.24$ & 61.58\small$\pm 0.87$ & 76.70\small$\pm 0.74$ & 91.35\small$\pm 0.42$ & 68.20\small$\pm 0.46$ & 93.37\small$\pm 0.32$ & 50.48\small$\pm 0.34$ & 60.32\small$\pm 0.65$ & 70.62\small$\pm 0.85$ & 77.64\small$\pm 0.20$ \\
        ANS-GT & 79.35\small$\pm 0.90$ & 64.52\small$\pm 0.71$ & 77.80\small$\pm 0.65$ & 80.41\small$\pm 0.78$ & 72.34\small$\pm 0.50$ & 95.30\small$\pm 0.81$ & - & - & 74.67\small$\pm 0.65$ & 80.64\small$\pm 0.29$\\
        \midrule
        HSGT & \textbf{83.56}\small$\pm 1.77$ & 67.41\small$\pm 0.92$ & \textbf{79.65}\small$\pm 0.52$ & \textbf{95.01}\small$\pm 0.34$ & \textbf{72.58}\small$\pm 0.31$ & \textbf{97.30}\small$\pm 0.24$ & \textbf{54.12}\small$\pm 0.51$ & \textbf{63.47}\small$\pm 0.45$ & \textbf{78.13}\small$\pm 0.25$ & \textbf{81.15}\small$\pm 0.13$\\
        \bottomrule
    \end{tabular}
    \end{adjustbox}
    \caption{Results on node classification datasets. OOM stands for out of memory. * indicates results cited from the original papers and the OGB leaderboard.}
    \label{results_1}
    \vspace{-6mm}
\end{table*}

\section{Experiments}

In this section we first evaluate HSGT on different benchmark tasks, and then perform ablation studies, scalability tests, and parameter analysis.
\subsection{Node Classification Tasks}

\paragraph{Datasets} To give thorough evaluation, we conduct experiments on nine benchmark datasets including four small-scale datasets (Cora, CiteSeer, PubMed \cite{sen2008collective,yang2016revisiting}, Amazon-Photo \cite{shchur2018pitfalls}) and six large-scale datasets (ogbn-arxiv, ogbn-proteins, ogbn-products \cite{hu2020open}, Reddit \cite{hamilton2018inductive}, Flickr, Yelp \cite{zeng2019graphsaint}). We use the predefined dataset split if possible, or we set a random \textbf{1:1:8} train/valid/test split.
\paragraph{Baselines and Settings}
We compare HSGT against a wide range of baseline scalable graph learning methods including GCN \cite{kipf2016semi}, GAT \cite{velivckovic2017graph}, GIN \cite{xu2018powerful}, GraphSAGE \cite{hamilton2018inductive}, Cluster-GCN \cite{chiang2019cluster}, GraphSAINT \cite{zeng2019graphsaint}, GAS-GCN \cite{fey2021gnnautoscale}, SIGN \cite{frasca2020sign}, GraphZoom \cite{deng2019graphzoom} and graph Transformers including Graphormer \cite{ying2021transformers}, SAN \cite{kreuzer2021rethinking}, SAT \cite{chen2022structure} and ANS-GT \cite{zhang2022hierarchical}. For GCN, GAT and GIN, we perform full-batch training on small-scale datasets and sampling-based training on large-scale datasets. By default, Graphormer, SAN and SAT require full-batch training, which is prohibited by GPU memory bound in most cases. We also add the Graphormer-\textsc{Sample} and SAT-\textsc{Sample} baselines that perform the model on subgraphs generated from neighborhood sampling, as mentioned in the introduction. For all experiments, the overhead of preprocessing steps (including METIS partition) does not exceed 5 minutes. The detailed settings for baselines and HSGT are listed in the appendix. We report the means and standard deviations of performances on test set initialized by three different random seeds, and code for implementing our method is provided in the supplementary material.

\paragraph{Results}
We present the node classification performances in Table \ref{results_1}, where the metric for \textsc{ogbn-proteins} is roc-auc while the metric for other datasets is acc. On small-scale datasets where graph Transformer baselines are mostly outperformed by GNN methods, HSGT delivers competitive performance against the GNN baselines. While on large-scale datasets, HSGT performs consistently better than all GNN baselines, achieving state-of-the-art and showing that the Transformer architecture is well capable of handling large-scale node-level tasks. Overall, the results have also demonstrated the outstanding generalizability of the HSGT method on multi-scale graph data.

It can also be observed that Transformers generally perform bad at node-level tasks on small graphs probably because the global attention introduces much irrelevant noise. And the Graphormer-\textsc{Sample} and SAT-\textsc{Sample} baseline fail to produce satisfactory results on multi-level benchmarks since a naive sampling-based approach can not capture the necessary high-level contextual information. On the contrary, on all datasets HSGT performs significantly better than the Graphormer-\textsc{Sample} and SAT-\textsc{Sample} baseline, showing the effectiveness of our proposed graph hierarchical structure and training strategies. Notably, the performance gains of HSGT are greater on large-scale datasets than small-scale ones, indicating that a large amount of data is crucial in optimizing the performance potentials of Transformer.

\subsection{Ablation Studies}


\paragraph{Settings.} We design four HSGT variants to demonstrate the benefits of vertical blocks, structural encodings, historical embeddings and readout blocks, respectively, while other model settings stay unchanged. In Table \ref{ablation_results}, \textit{w/o vertical blocks}: the vertical feature aggregation is performed with the simple mean function, instead of a Transformer block. \textit{w/o structural encodings}: all SPD attention biases are removed. \textit{w/o historical embeddings}: neighborhood nodes of high-level nodes are no longer sampled, then historical embeddings are no longer needed. \textit{w/o readout blocks}: the multi-level readout is performed by concatenating feature vectors at different levels. \textit{w/o parameter sharing}: all horizontal blocks and vertical blocks have an individual set of parameters.  \textit{random partition}: the coarsening process is performed via random partition, instead of METIS algorithm.

\begin{table}[h]
    \begin{adjustbox}{width=0.9\columnwidth,center}
    \begin{tabular}{l|ccccc}
        \toprule
        & \textsc{Flickr} & \textsc{Yelp} & \texttt{ogbn-products} \\
        \midrule
        \textit{w/o vertical blocks} & 52.85 & 62.46 & 78.01 \\
        \textit{w/o structural encodings} & 49.11 & 59.78 & 77.05\\
        \textit{w/o historical embeddings} & 52.04 & 61.79 & 80.72\\
        \textit{w/o readout blocks} & 52.58 & 62.01 & 80.34\\
        \textit{w/o parameter sharing} & 53.01 & 63.44 & 81.13 \\
        \textit{random partition} & 43.21 & 60.77 & 75.23 \\
        \midrule
        HSGT & \textbf{53.02} & \textbf{63.47} & \textbf{81.15}\\
        \bottomrule
    \end{tabular}
    \end{adjustbox}
    \caption{Results of ablation studies.}
    \label{ablation_results}
    \vspace{-4mm}
\end{table}

\begin{table*}[t]
    \begin{adjustbox}{width=1.6\columnwidth,center}
    \begin{tabular}{l|ccccc}
        \toprule
        Model & Settings & Peak GPU Memory Usage & \#Parameters & Inference Time & Performance \\
        \midrule
        \multirow{4}{*}{GraphSAGE} & $l=2,d=64$ & 1064MB & 16.4K & 48.6s & 75.60 \\
        & $l=2,d=128$ & 1150MB & 65.5K & 53.7s & 77.38 \\
        & $l=2,d=256$ & 1316MB & 262.1K & 70.2s & 77.58 \\
        & $l=3,d=128$ & 1928MB & 98.3K & 103.3s & 78.25 \\
        \midrule
        \multirow{2}{*}{Graphormer-\textsc{Sample}} & $l=2,d=128$ & 1874MB & 223.6K & 90.6s & 63.44 \\
        & $l=2,d=256$ & 2032MB & 840.2K & 93.5s & 64.76 \\
        \midrule
        \multirow{3}{*}{HSGT} & $l=2,d=64$ & 1903MB & 112.7K & 97.4s & 80.99 \\
        & $l=2,d=128$ & 2208MB & 421.1K & 99.8s & 81.10 \\
        & $l=3,d=64$ & 3374MB & 137.7K & 108.2s & \textbf{81.15} \\
        \bottomrule
    \end{tabular}
    \end{adjustbox}
    \caption{Results of scalability tests on \texttt{ogbn-products} dataset.}
    \label{scale_results}
    \vspace{-3mm}
\end{table*}

\paragraph{Results.} Table \ref{ablation_results} summarizes the results of ablation studies. Most variants suffer from performance loss, showing that all tested modules are necessary to raise the performance of HSGT to its best level. If we replace Transformer layer in vertical and readout blocks with simple operations like mean, then multi-scale information can not be adaptively fused. And the model will not be able to recognize graph structure when we remove the necessary structural encodings, which explains the severe performance drop we witness. It can also be observed that if we take out the neighborhood sampling of high-level nodes to avoid historical embeddings, the high-level context information exchange can be blocked, resulting in performance drop on large-scale datasets. When individual learnable parameters are assigned to horizontal blocks at different levels, the model performance is almost not affected while the number of parameters could increase by at least $1.5\times$. A random coarsening approach also causes the model performance to drop dramatically because HSGT requires high-quality structural partitions calculated by the METIS algorithm to effectively capture and aggregate high-level information. Those results can well confirm our previous statements.

\subsection{HSGT Efficiently Scales to Large Graphs}
\paragraph{Settings.} To comprehensively examine HSGT's scalability to large-scale graphs, we perform tests on the largest \texttt{ogbn-products} dataset that contains over 2.4 million nodes against the standard GraphSAGE method and Graphormer-\textsc{Sample} above. To give a fair comparison, for GraphSAGE and Graphormer-\textsc{Sample} we set the batch size to 400, and for HSGT we keep layer 0 nodes per batch to around 400. In Table \ref{scale_results}, $l$ stands for the number of layers for GraphSAGE and Graphormer-\textsc{Sample}, while number of Transformer layers at horizontal blocks for HSGT. $d$ stands for the hidden dimension for all models. Other model parameters stay the same with experiments in Table \ref{results_1}. We use built-in PyTorch CUDA tools to measure peak GPU memory usage during experiments. For the number of parameters, we calculate the number of all learnable model parameters except input and output projections. For inference time, we measure the time for conducting inference on the whole dataset using PyTorch performance counters. 
\paragraph{Results.} In Table \ref{scale_results} we list the results. Traditionally, it is believed that the Transformer architecture tends to achieve high performance with high computational complexity and lots of parameters. However, experimental results show that the proposed HSGT model can achieve outstanding performance with reasonable costs that are similar to the widely-used GraphSAGE method, showing that HSGT can be efficiently scaled to large graphs with normal computational resources. Even under the lightest setting $l=2,d=64$, HSGT is capable of delivering results higher than all GNN baselines while keeping moderate GPU memory usage and parameter size, which can be attributed to the small hidden size (64) and the parameter sharing among horizontal blocks. The Graphormer-\textsc{Sample} may cost fewer resources at light-weight configurations since HSGT has the additional horizontal and vertical blocks, but it is significantly outperformed by both GraphSAGE and HSGT.

\subsection{Parameter Analysis}
\subsubsection{Coarsening Ratios} The coarsening ratios $\alpha_1,\ldots,\alpha_H$ are used as parameters for the METIS algorithm to generate initial graph hierarchy $\{G^0,G^1,\ldots,G^H\}$. Normally we set the number of additional hierarchical layers $H$ to 1 or 2, and coarsening ratios are picked from $\{0.2,0.1,0.05,0.02,0.01,0.005\}$. Other settings stay the same with models in Table \ref{results_1}. Here we perform experiments on dataset Flickr, Yelp and ogbn-products to study the influence of coarsening ratios, and we list the results in Table \ref{coarsening_results}. It can be observed that the best setting for coarsening ratios and batch size could vary for different datasets.
\begin{table}[h]
    \begin{adjustbox}{width=0.8\columnwidth,center}
    \begin{tabular}{l|ccc}
        \toprule
        Coarsening Ratios & \textsc{Flickr} & \textsc{Yelp} & \texttt{ogbn-products} \\
        \midrule
        $\{0.005\}$ & \textbf{52.71} & 61.98 & \textbf{81.15} \\
        $\{0.002\}$ & 51.32 & 61.44 & 80.55\\
        $\{0.1,0.1\}$ & 47.59 & 62.95 & OOM\\
        $\{0.1,0.2\}$ & 47.03 & \textbf{63.47} & OOM\\
        \bottomrule
    \end{tabular}
    \end{adjustbox}
    \caption{Results of coarsening ratio tests.}
    \label{coarsening_results}
    \vspace{-4mm}
\end{table}

 
\subsubsection{Intra-batch Connectivity}
At previous sections we have mentioned that in horizontal blocks, we construct the receptive field of each node with its $D$-hop neighbors and nodes randomly sampled from the sampled batch by probability $p$. Here we study the impact of $p$ value with experiments and summarize the results in Table \ref{p_results}, where other settings stay the same with those in Table \ref{results_1}. From the results we can see that as $p$ varies from $0$ to $1$, the model performance generally increases until it reaches a peak and then decreases, which corresponds to our previous analysis.

\begin{table}[h]
    \begin{adjustbox}{width=0.8\columnwidth,center}
    \begin{tabular}{c|ccc}
        \toprule
        Intra-batch Connectivity & \textsc{Flickr} & \textsc{Yelp} & \texttt{ogbn-products} \\
        \midrule
        $p=0.0$ & 51.04 & 62.83 & 80.45 \\
        $p=0.1$ & 50.80 & \textbf{63.47} & \textbf{81.15} \\
        $p=0.3$ & \textbf{53.02} & 62.92 & 80.67 \\
        $p=0.5$ & 48.47 & 60.41 & 80.14 \\
        $p=1.0$ & 45.83 & 60.74 & 78.65 \\
        \bottomrule
    \end{tabular}
    \end{adjustbox}
    \caption{Results of intra-batch connectivity tests.}
    \label{p_results}
    \vspace{-4mm}
\end{table}

\section{Conclusion}
In this paper we propose HSGT, a completely Transformer-based neural architecture for scalable graph learning, and our model has shown strong performance and generalizability on multi-scale graph benchmarks with reasonable computational costs.

\clearpage

\section*{Acknowledgements}
This work was supported by the National Natural Science Foundation of China (Grant No. 62276006).

\bibliographystyle{named}
\bibliography{main}

\clearpage

\appendix

\section{More Discussion}

\subsection{Hierarchy Construction Algorithm} One key component in HSGT is the graph hierarchy construction algorithm. In principle, our model can work on any graph hierarchical map, including randomly generated ones. However, considering in HSGT high-level nodes are mainly designed to expand the receptive field of low-level nodes by aggregating and transferring structural knowledge in low-level substructures, it is reasonable to expect that coarsening methods with better preservation of local topology can lead to higher performance. Besides, the method must be computationally efficient to work on large-scale graphs, which prohibits the usage of adaptive pooling methods like DiffPool \cite{diffpool}. In this sense, we mainly use METIS \cite{karypis1997metis} which is the prevalent method that accurately partitions the topological structure of graphs and is scalable to graphs of millions of nodes (up to 5 minutes in our tested datasets). Another key advantage of METIS is that its coarsening ratio can be arbitrarily set, so we can test on different ratios and optimize the performance and computational cost of the model, which leads us to choose METIS instead of the famous Graclus \cite{dhillon2007weighted}.

\subsection{Comparison Against GNN-based Methods} Many existing GNN-based methods like \cite{diffpool,deng2019graphzoom} have proposed to enhance the expressivity and scalability of GNNs via graph hierarchies. However, comparing with these methods, HSGT is fully Transformer-based and utilizes the graph hierarchical structure from a different motivation. GNN-based methods generally use the coarsened graph to improve the embedding quality by summarizing the graph topological structure and provide global information to node-level tasks. While for HSGT, we utilize the graph hierarchical structure to efficiently expand the receptive field of low-level nodes. Avoiding directly scaling up the sampling batch size, we store high-level context information with representations of high-level virtual nodes, and through multi-level attention the receptive field of nodes is effectively expanded.

\section{Experiment Settings}

The experiment code is available at \url{https://drive.google.com/file/d/1HI5HYRdrv-B-o6HSECpwLxs9vq6DsjPo/view?usp=share_link}. We build the HSGT model with 2 Transformer layers for every horizontal block and the number of attention heads is fixed to 8. The maximum SPD length $D$ in every horizontal block is set to 2. We apply the AdamW optimizer and set the weight decay value to 1e-5. In subgraph sampling, we sample 5 neighbors and 10 2-hop neighbors for every layer 0 node, and only sample 5 neighbors for high-level nodes. The search space for hidden dimension, dropout rate, learning rate, random intra-batch sampling ratio are $\{64,128,256\}$,$\{0.1,0.3,0.5\},\{1e-3,3e-4,1e-4\},\{0.0,0.1,0.3\}$, respectively. All experiments are performed on a machine with one NVIDIA A30 GPU, and the training may take up to 6 hours. We present the parameter settings for HSGT node classification experiments in Table \ref{small_params} and \ref{large_params}.

For all baseline methods, we conduct hyper-parameter search and report the final test set performance. For GCN, GIN, GAT and GraphSAGE, we set the batch size to 512, dropout rate to 0.5, and the search places for learning rate, number of GNN layers, hidden size are $\{0.001,0.005,0.01,0.1\},\{2,3,4\},\{64,128,256\}$, respectively. For GAT, the number of attention heads is set to 4. For Cluster-GCN, except the parameters above, we find the best number of partitions from $\{150,1500,15000\}$ and set the batch size to 20. For GraphSAINT, except the parameters for GCN, GIN and GAT, we set the dropout rate to 0.2, sampling random walk length to 2, number of iterations per epoch to 5, and number of samples to compute normalization statistics to 100. For SGC, the number of propagations is searched from $\{2,3,4\}$. For Graphormer and Graphormer-\textsc{Sample}, the search spaces for number of Transformer layers and hidden size are $\{2,3,4\}$ and $\{64,128,256\}$. For Graphormer-\textsc{Sample}, we search the batch size from $\{200,400,600\}$.

\begin{table*}[t]
    \begin{adjustbox}{width=1.5\columnwidth,center}
    \begin{tabular}{l|cccc}
    \toprule
    Dataset & \textsc{Cora} & \textsc{CiteSeer} & \textsc{PubMed} & \textsc{Amazon-Photo}\\
    \midrule
    Coarsening Ratios & $\{0.05\}$ & $\{0.05\}$ & $\{0.05\}$ & $\{0.05\}$ \\
    Batch Size & full-batch & 400 & 100 & 100 \\
    Hidden Size & 256 & 256 & 128 & 256 \\
    Number of Transformer Layers in Horizontal Block & 3 & 3 & 3 & 3 \\
    \bottomrule
    \end{tabular}
    \end{adjustbox}
    \caption{HSGT Settings on small-scale Datasets.}
    \label{small_params}
\end{table*}

\begin{table*}[t]
    \begin{adjustbox}{width=1.5\columnwidth,center}
    \begin{tabular}{l|cccccccccc}
    \toprule
    Dataset & \texttt{ogbn-arxiv} & \textsc{Reddit} & \textsc{Flickr} & \textsc{Yelp} & \texttt{ogbn-proteins} & \texttt{ogbn-products} \\
    \midrule
    Coarsening Ratios & $\{0.01\}$ & $\{0.005\}$ & $\{0.01\}$ & $\{0.1,0.2\}$ & $\{0.005\}$ & $\{0.005\}$ \\
    Batch Size & 4 & 4 & 4 & 2 & 2 & 4 \\
    Hidden Size & 256 & 128 & 64 & & 64 64 & 64 \\
    Number of Transformer Layers in Horizontal Block & 2 & 3 & 3 & 2 & 3 & 3 \\
    \bottomrule
    \end{tabular}
    \end{adjustbox}
    \caption{HSGT Settings on large-scale Datasets.}
    \label{large_params}
\end{table*}


\end{document}